\documentclass[10pt,twocolumn,letterpaper]{article}

\usepackage{cvpr}              

\usepackage[dvipsnames,table,svgnames]{xcolor}

\usepackage{comment}
\usepackage{multirow}
\usepackage{pifont}
\usepackage{makecell}
\usepackage{booktabs,siunitx}

\definecolor{Gray}{gray}{0.85}
\definecolor{whitesmoke}{rgb}{0.96, 0.96, 0.96}
\definecolor{LightCyan}{rgb}{0.88,1,1}
\newcolumntype{g}{>{\columncolor{LightCyan}}c}

\definecolor{cvprblue}{rgb}{0.21,0.49,0.74}
\usepackage[pagebackref,breaklinks,colorlinks,citecolor=cvprblue]{hyperref}

\def\paperID{10435} 
\def\confName{CVPR}
\def\confYear{2024}

\title{Self-Supervised Facial Representation Learning with Facial Region Awareness}

\author{Zheng Gao \and Ioannis Patras \and \\
	Queen Mary University of London, Mile End Road, London, E1 4NS \\
	{\tt\small \{z.gao, i.patras\}@qmul.ac.uk}
}

\begin{document}
\maketitle

\begin{abstract}
	Self-supervised pre-training has been proved to be effective in learning transferable representations that benefit various visual tasks. This paper asks this question: can self-supervised pre-training learn general facial representations for various facial analysis tasks? Recent efforts toward this goal are limited to treating each face image as a whole, \ie, learning consistent facial representations at the image-level, which overlooks the ``\textbf{consistency of local facial representations}'' (\ie, facial regions like eyes, nose, etc). In this work, we make a \textbf{first attempt} to propose a novel self-supervised facial representation learning framework to learn consistent global and local facial representations, \textbf{F}acial \textbf{R}egion \textbf{A}wareness (FRA). Specifically, we explicitly enforce the consistency of facial regions by matching the local facial representations across views, which are extracted with learned heatmaps highlighting the facial regions. Inspired by the mask prediction in supervised semantic segmentation, we obtain the heatmaps via cosine similarity between the per-pixel projection of feature maps and ``facial mask embeddings'' computed from learnable positional embeddings, which leverage the attention mechanism to globally look up the facial image for facial regions. To learn such heatmaps, we formulate the learning of facial mask embeddings as a deep clustering problem by assigning the pixel features from the feature maps to them. The transfer learning results on facial classification and regression tasks show that our FRA outperforms previous pre-trained models and more importantly, using ResNet as the unified backbone for various tasks, our FRA achieves comparable or even better performance compared with SOTA methods in facial analysis tasks.
\end{abstract}

\section{Introduction}
Human face understanding is an important and challenging topic in computer vision~\cite{zheng2022general,liu2023pose} and supervised learning algorithms have shown promising results on a wide range of facial analysis tasks recently~\cite{Zhao_2016_CVPR,cao2018partially,kumar2020luvli,te2021agrnet}. Despite the impressive progress, these methods require large-scale well-annotated training data, which is expensive to collect.

Recent works in self-supervised representation learning for visual images have shown that self-supervised pre-training is effective in improving the performance on various downstream tasks such as image classification, object detection and segmentation as it can learn general representations from unlabeled data that could be transferred to downstream visual tasks, especially tasks with limited labeled data~\cite{ji2019invariant,khosla2020supervised,cvpr19unsupervised,hjelm2019learning,dwibedi2021little,wang2021dense,xie2021propagate}. Among them, \textbf{instance discrimination} (including contrastive learning~\cite{pmlr-v119-chen20j,he2020momentum,li2021prototypical} and non-contrastive learning~\cite{grill2020bootstrap,chen2020exploring} paradigms) has been shown to be effective in learning generalizable self-supervised features. Instance discrimination aims to learn view-invariant representations by matching the \textbf{global representations} between the augmented views generated by aggressive image augmentations, \ie, the image-level representations of the augmented views should be similar~\cite{pmlr-v119-chen20j,he2020momentum,chen2020mocov2, li2021prototypical,grill2020bootstrap,chen2020exploring}. Another self-supervised learning paradigm, masked image modeling (MIM)~\cite{he2022masked,tao2023siamese} learns visual representations by reconstructing image content from a masked image, achieving excellent performance in full model fine-tuning. This leads to the question: \textbf{can self-supervised pre-training learn general facial representations which benefit downstream facial analysis tasks}?

Several attempts have been made to learn general facial representations for facial analysis tasks~\cite{bulat2022pre,zheng2022general,liu2023pose}. For example, Bulat \etal~\cite{bulat2022pre} directly applies the contrastive objective to facial features. FaRL~\cite{zheng2022general} and MCF~\cite{wang2023toward} combine contrastive learning and mask image modeling~\cite{he2022masked}. PCL~\cite{liu2023pose} proposes to disentangle the pose-related and pose-unrelated features, achieving strong performance on both pose-related (regression) and pose-unrelated (classification) tasks. However, it runs the model forward and backward three times for each training step, which is time-consuming. Despite different techniques, these methods commonly treats each face image as a whole to learn consistent global representations at the image-level and overlook the ``\textbf{spatial consistency of local representations}'', \ie, local facial regions (\eg, eyes, nose and mouth) should also be similar across the augmented views, thus limiting the generalization to downstream tasks. This brings us to the focus: \textbf{learning consistent global and local representations for facial representation learning}.

We argue that in order to learn consistent local representations, the model needs to look into facial regions. Towards that goal, we predict a set of heatmaps highlighting different facial regions by leveraging learnable positional embeddings as facial queries (the feature maps as keys and values) to look up the facial image globally for facial regions, which is inspired by the mask prediction in supervised segmentation~\cite{cheng2021per}. For visual images, the attention mechanism of Transformer allows the learnable positional embeddings to serve as object queries for visual pattern look-up~\cite{carion2020end,cheng2021per}. In our case (facial images), the learnable positional embeddings can be used as facial queries for facial regions (\textbf{see the visualization in the supplementary material}).

In this work, taking the consistency of facial regions into account, we make a \textbf{first attempt} to propose a novel self-supervised facial representation learning framework, \textbf{F}acial \textbf{R}egion \textbf{A}wareness (FRA) that learns general facial representations by enforcing consistent global and local facial representations, based on a popular instance discrimination baseline BYOL~\cite{grill2020bootstrap} for its simplicity. Specifically, we \textbf{learn consistent local facial representations} by match them across augmented views, which are extracted by aggregating the feature maps using learned heatmaps highlighting the facial regions as weights. Inspired by the mask prediction in MaskFormer~\cite{cheng2021per}, we produce the heatmaps from a set of learnable positional embeddings, which are used as facial queries to look up the facial image for facial regions. A Transformer decoder takes as input the feature maps from the encoder and the learnable positional embeddings to output a set of ``\textit{facial mask embeddings}'', each associated with a facial region. The facial mask embeddings are used to compute cosine similarity with the per-pixel projection of feature maps to produce the heatmaps. In addition, we enforce the consistency of global representations across views simultaneously so that the image-level information is preserved. In order to learn the heatmaps (facial mask embeddings), inspired by deep clustering~\cite{caron2020unsupervised} that learns to assign samples to clusters, we treat the facial mask embeddings as clusters and learn to assign pixel features from the feature maps to them. Specifically, we \textbf{align the per-pixel cluster assignments} of each pixel feature over the facial region clusters between the online and momentum network for the same augmented view (\ie, each pixel feature should have similar similarity distribution over the facial mask embeddings between the momentum teacher and online student). In contrast to supervised segmentation that directly uses ground-truth masks to supervise the learning of the masks (heatmaps) with a per-pixel binary mask loss, we formulate the learning of heatmaps as a deep clustering~\cite{caron2020unsupervised} problem that learns to assign pixel features to clusters (facial mask embeddings) in a self-supervised manner.

Our contributions can be summarized as follows:
\begin{itemize}
	\item Taking into the consistency of local facial regions into account, we make a \textbf{first attempt} to propose a novel self-supervised facial representation learning framework, \textbf{F}acial \textbf{R}egion \textbf{A}wareness (FRA) that learns consistent global and local facial representations.
	\item We show that the learned heatmaps can roughly discover facial regions in the supplementary material.
	\item In previous works, different backbones are adopted for different facial analysis tasks (\eg, in face alignment the common backbone is hourglass network~\cite{yang2017stacked} while in facial expression recognition ResNet~\cite{he2016deep} is the popular backbone). In this work, our FRA achieves SOTA results using vanilla ResNet~\cite{he2016deep} as the unified backbone for various facial analysis tasks.
	\item Our FRA outperforms existing self-supervised pre-training approaches (\eg, BYOL~\cite{grill2020bootstrap} and PCL~\cite{liu2023pose}) on facial classification (\ie, facial expression recognition~\cite{goodfellow2013challenges,li2017reliable} and facial attribute recognition~\cite{liu2015deep}) and regression (\ie, face alignment~\cite{wu2018look,sagonas2016300,sagonas2013300,sagonas2013semi}) tasks. More importantly, our FRA achieves comparable (\eg, face alignment) or even better performance (\eg, facial expression recognition) compared with SOTA methods in the corresponding facial analysis tasks.
\end{itemize}

\section{Related work}
\subsection{Visual representation learning}
As one of the main paradigms for self-supervised pre-training, instance discrimination learns representations by treating an image as a whole and enforcing the consistency of global representations at the image-level across augmented views. Generally, instance discrimination includes two paradigms: contrastive learning~\cite{pmlr-v119-chen20j,he2020momentum,li2021prototypical} and non-contrastive learning ~\cite{grill2020bootstrap,chen2020exploring}. Contrastive learning considers each image and its transformations as a distinct class, \ie, ``positive'' samples are pulled together while ``negative'' samples are pushed apart in the latent space. Unlike contrastive learning that relies on negative samples to avoid collapse, non-contrastive learning directly maximizes the similarity of the global representations between the augmented views without involving negative samples based on techniques like stop-gradient~\cite{chen2020exploring} and predictor~\cite{grill2020bootstrap}. Further works perform visual-language pre-training by applying contrastive objective to image-text pairs~\cite{radford2021learning,jia2021scaling,li2021align}.

Another line of work, masked image modeling (MIM) learns visual representations by reconstructing image content from a masked image~\cite{he2022masked,assran2022masked,tao2023siamese}, which is inspired by the masked language modeling in NLP~\cite{devlin2018bert}. In contrast to instance discrimination, MIM achieves strong full model fine-tuning performance with Vision Transformers pre-trained for enough epochs. However, these works suffer from poor linear separability and are less data-efficient than instance discrimination in few-shot scenarios~\cite{assran2022masked}.

\subsection{Facial representation learning}
Recent works on facial analysis explore self-supervised learning for several face-related tasks, such as facial expression recognition~\cite{chang2021learning,Shu_2022_BMVC}, face recognition~\cite{chang2021learning,wang2023ucol}, facial micro-expression recognition~\cite{nguyen2023micron}, AU detection~\cite{li2020learning,li2019self}, face alignment (facial landmark detection)~\cite{cheng2021equivariant,yang2022dense}, etc. However, these methods are \textbf{task-specific}, \ie, tailored for a specific facial task and thus lack the ability to generalize to various facial analysis tasks~\cite{liu2023pose}. Further efforts~\cite{bulat2022pre,zheng2022general,liu2023pose} focus on learning general facial representations with contrastive learning and mask image modeling~\cite{he2022masked,tao2023siamese}. Bulat \etal~\cite{bulat2022pre} directly apply the contrastive objective to augmented views of the same face image, showing that general facial representation learned from pre-training can benefit various facial analysis tasks. FaRL~\cite{zheng2022general} performs pre-training in a visual-linguistic manner by employing image-text contrastive learning and masked image modeling. MCF~\cite{wang2023toward} leverages image-level contrastive learning and masked image modeling, along with the knowledge distilled from external ImageNet pre-trained model for facial representation learning. PCL~\cite{liu2023pose} argues that directly applying the contrastive objective to face images overlooks the variances of facial poses and thus leads to pose-invariant representations, limiting the performance on pose-related tasks~\cite{zhu2016face,yin20063d}. Therefore, PCL~\cite{liu2023pose} disentangles the pose-related and pose-unrelated features and then performs contrastive learning on these features, achieving strong performance on both pose-related and pose-unrelated facial analysis tasks. Despite the success, it performs forward and backward three times for each input image, which brings significant increase on computational cost. These works are commonly limited by instance discrimination and overlook the consistency of local facial regions. In contrast, inspired by supervised semantic segmentation, we learn consistent global and local facial representations by learning a set of heatmaps indicating facial regions from learnable positional embeddings, which leverage the attention mechanism to look up facial image globally for facial regions.

\subsection{Facial region discovery}
There are some approaches leveraging facial region (landmark) discovery for facial analysis~\cite{jakab2018unsupervised,xia2022sparse,Lu_2023_CVPR}. Some focus on landmark detection by either learning a heatmap for each landmark via image reconstruction~\cite{jakab2018unsupervised,zhang2018unsupervised}, or performing pixel-level matching with an equivariance loss~\cite{thewlis2019unsupervised,zhang2018unsupervised}. Despite different techniques, these methods are \textbf{task-specific}, \ie, landmark detection with discovery of local information, while our method is \textbf{task-agnostic}, \ie, learn general facial representations for various tasks by preserving global and local information in a \textbf{image, region and pixel-level contrastive manner}. MARLIN~\cite{cai2023marlin} applies masked image modeling to learn general representations for facial videos by utilizing an external face parsing algorithm to discover the facial regions (\eg, eyes, nose and mouth), which are used to guide the masking for the masked autoencoder. A closely related work SLPT~\cite{xia2022sparse} leverages the attention mechanism to estimate facial landmarks from initial facial landmarks estimates of the mean face through supervised learning. These works commonly rely on external supervisory signal, whether it is from ground-truths or additional algorithms. In contrast, we learn to discover the facial regions in an \textbf{end-to-end self-supervised} manner for facial \textbf{image} representation learning.

\begin{figure*}[htb]
	\centering
	\includegraphics[width=0.9\linewidth]{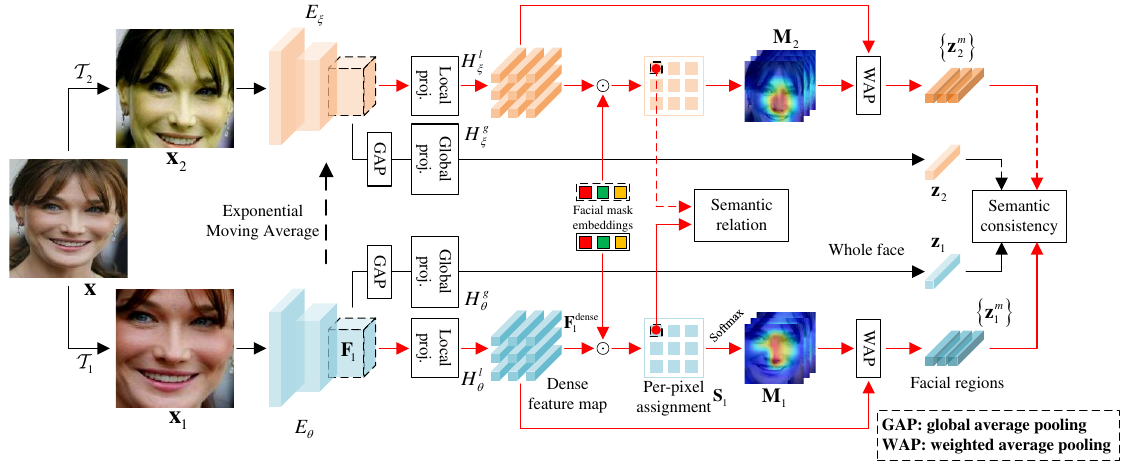}
	\caption{\textbf{Overview of the proposed FRA framework}. $\odot$ denotes cosine similarity. For each input image $\mathbf{x}$, its augmented views $\mathbf{x}_1$ and $\mathbf{x}_2$ are passed into two network branches to produce the global embeddings $\mathbf{z}_1$ and $\mathbf{z}_2$. In addition, we produce a set of heatmaps $\mathbf{M}_1$ and $\mathbf{M}_2$ indicating the local facial regions, via the correlation between the pixel features and ``\textit{facial mask embeddings}'' computed from a set of learnable positional embeddings. Then we aggregate the feature map to obtain the local facial embeddings $\{\mathbf{z}^{m}_1\}$ and $\{\mathbf{z}^{m}_2\}$. The semantic consistency loss is applied to global embeddings and facial embeddings to maximize the similarity across augmented views. To learn such heatmaps, \ie, \textit{facial mask embeddings}, we treat the facial mask embeddings as facial region clusters and propose a semantic relation loss to align the cluster assignments of each pixel feature over the facial region clusters between the online and momentum network.}
	\label{fig:FRA-overview}
\end{figure*}

\begin{figure}[htb]
	\centering
	\includegraphics[width=\linewidth]{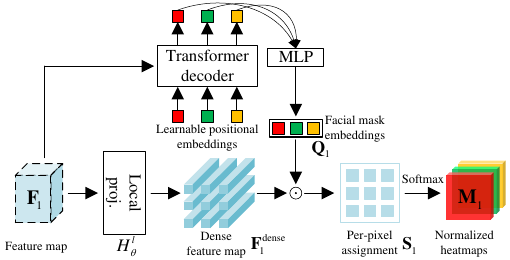}
	\caption{Generation of heatmaps using learnable positional embeddings as facial queries and the feature maps as keys and values.}
	\label{fig:FRA-transformer}
\end{figure}

\section{Methodology}
\subsection{Overview}\label{sec:FRA-overview}
The overview of the proposed FRA is shown in~\cref{fig:FRA-overview}. The goal is to learn consistent global and local facial representations. Toward this goal, we propose two objectives: \textbf{pixel-level semantic relation} and \textbf{image/region-level semantic consistency}. Semantic relation aligns the per-pixel cluster assignments of each pixel feature over the facial mask embeddings between the online and momentum network to learn heatmaps for facial regions (\cref{sec:FRA-relation}) while semantic consistency directly matches the global and local facial representations across augmented views (\cref{sec:FRA-consistency}) with the learned heatmaps.

\subsection{Semantic relation}\label{sec:FRA-relation}
As shown in~\cref{fig:FRA-overview}, our method adopts the Siamese structure of BYOL~\cite{grill2020bootstrap}, a popular self-supervised pre-training baseline based on instance discrimination. Following BYOL~\cite{grill2020bootstrap}, we employ two branches: the online network parameterized by $\theta$ and the momentum network parameterized by $\xi$. The online network $\theta$ consist of an encoder $E_\theta$, a global projector $H^g_\theta$ and a local projector $H^l_\theta$. The momentum network $\xi$ has the same architecture with the online network, except $\xi$ is updated with an exponential moving average of $\theta$. As in BYOL~\cite{grill2020bootstrap}, we also adopt additional predictors $G^g_\theta$ and $G^l_\theta$ on top of the projectors in the online network. Note that this is omitted for brevity in~\cref{fig:FRA-overview}.

Given an input image $\mathbf{x}$, two random augmentations are applied to generate two augmented views $\mathbf{x}_1=\mathcal{T}_1(\mathbf{x})$ and $\mathbf{x}_2=\mathcal{T}_2(\mathbf{x})$, following BYOL~\cite{grill2020bootstrap}. Each augmented view $\mathbf{x}_i \in \{\mathbf{x}_1, \mathbf{x}_2\}$ is fed into an encoder $E$ to obtain a feature map $\mathbf{F}_i \in \mathbb{R}^{C\times H\times W}$ (before global average pooling), where $C$, $H$, $W$ are the number of channels, height and width of $\mathbf{F}_i$ and a latent representation $\mathbf{h}_i \in \{\mathbf{h}_1, \mathbf{h}_2\}$ (after global average pooling), \ie, $\mathbf{h}_1 = E_\theta(\mathbf{x}_1)$ and $\mathbf{h}_2 = E_\xi(\mathbf{x}_2)$. Then each latent representation $\mathbf{h}_i$ is transformed by a global projector $H^g$ to produce a global embedding $\mathbf{z}_i \in \{\mathbf{z}_1, \mathbf{z}_2\}$ of dimension $\mathbf{z}_i \in \mathbb{R}^D$, \ie, $\mathbf{z}_1 = H^g_\theta(\mathbf{h}_1)$ and $\mathbf{z}_2 = H^g_\xi(\mathbf{h}_2)$.

Then we obtain a set of heatmaps $\mathbf{M}_i \in \{\mathbf{M}_1, \mathbf{M}_2\}$ highlighting the facial regions from the feature map $\mathbf{F}_i$ for each view, which is inspired by \textit{mask classification-based supervised segmentation}~\cite{carion2020end,cheng2021per} that leverages attention mechanism to look up visual patterns globally. First the local projector (\eg, $H^l_\theta$) is applied to project the pixel features of $\mathbf{F}_i$ in a pixel-wise manner, mapping it to $D$ dimensions to get the dense feature map $\mathbf{F}^\text{dense}_i \in \mathbb{R}^{D\times H\times W}$. Take view $\mathbf{x}_1$ as an example, the projected feature map can be expressed as:
\begin{equation}\label{eq:FRA-couple}
	\mathbf{F}^\text{dense}_1[*, u, v] = H^l_\theta(\mathbf{F}_1[*, u, v]),
\end{equation}
where $\mathbf{F}_1[*, u, v] \in \mathbb{R}^C$ is the pixel feature at the $(u,v)$-th grid of $\mathbf{F}_1$. Then as shown in~\cref{fig:FRA-transformer}, inspired by supervised segmentation~\cite{cheng2021per}, we use a Transformer decoder followed by a MLP, which takes as input the feature map $\mathbf{F}_i$ and $N$ learnable positional embeddings (\ie, facial queries for looking up the facial image globally for facial regions) to predict $N$ ``\textit{facial mask embeddings}'' $\mathbf{Q}_i \in \mathbb{R}^{N \times D}$ of dimension $D$, where each row associated with a facial region. Next, we compute the cosine similarity between facial mask embeddings $\mathbf{Q}_i$ and dense feature map $\mathbf{F}^\text{dense}_i$ along the channel dimension, yielding \textbf{per-pixel cluster assignments} $\mathbf{S}_i \in \mathbb{R}^{N \times H \times W}$, where $\mathbf{S}_i[*, u, v]$ denotes the relation between the dense pixel feature $\mathbf{F}^\text{dense}_1[*, u, v]$ and facial mask embeddings $\mathbf{Q}_i$. Finally, we normalize $\mathbf{S}_i$ along the channel dimension with a softmax operation to encourage each channel to capture a different pattern, obtaining $N$ heatmaps $\mathbf{M}_i \in \mathbb{R}^{N \times H \times W}$ where each vector at location $(u, v)$ is a probabilistic similarity distribution (\ie, \textbf{normalized per-pixel cluster assignments}) $\mathbf{s}^{u,v}_1$ between $\mathbf{F}^\text{dense}_i[*, u, v]$ and $\mathbf{Q}_i$. Note that $\mathbf{M}_i$ is a set of heatmaps where each channel of $\mathbf{M}_i$ represents a 2D heatmap $\mathbf{M}^{(m)}_i$.

To learn such heatmaps, \ie, facial mask embeddings, inspired by deep clustering~\cite{caron2020unsupervised}, we treat the facial mask embeddings as facial region clusters and align the per-pixel cluster assignments of each pixel feature over these clusters between the online and momentum network for the same augmented view, using the momentum network as momentum teacher~\cite{zhou2022image,dong2022bootstrapped} to provide reliable target.

Following BYOL~\cite{grill2020bootstrap}, we pass both augmented views $\mathbf{x}_1$ and $\mathbf{x}_2$ through the online and momentum network. Take $\mathbf{x}_1$ as an example, the online network $\theta$ outputs the normalized per-pixel cluster assignments $\mathbf{s}^{u,v}_1$ and the momentum network outputs normalized assignments $\mathbf{\widehat{s}}^{u,v}_1$ for view $\mathbf{x}_1$. Then we learn $\mathbf{s}^{u,v}_1$ using $\mathbf{\widehat{s}}^{u,v}_1$ as guidance based on the following cross-entropy loss:
\begin{equation}\label{eq:FRA-ce}
	CE(\mathbf{s}^{u,v}_1, \mathbf{\widehat{s}}^{u,v}_1) = -\sum_{m=1}^{N}{\mathbf{\widehat{s}}^{u,v}_1[m] \log{\mathbf{s}^{u,v}_1[m]}}.
\end{equation}

For both augmented views, we define the symmetrized semantic relation objective as:
\begin{equation}\label{eq:FRA-relation}
	\mathcal{L}_\text{r} = \frac{1}{HW}\sum_{u,v}{(CE(\mathbf{s}^{u,v}_1, \mathbf{\widehat{s}}^{u,v}_1) + CE(\mathbf{s}^{u,v}_2, \mathbf{\widehat{s}}^{u,v}_2))},
\end{equation}
where $CE(\mathbf{s}^{u,v}_2, \mathbf{\widehat{s}}^{u,v}_2)$ is the cross-entropy loss for view $\mathbf{x}_2$. We apply the Sinkhorn-Knopp normalization to the target assignments from the momentum network following~\cite{caron2020unsupervised} to avoid collapse and the mean entropy maximization (ME-MAX) regularizer~\cite{assran2022masked} to maximize the entropy of the prediction to encourage full use of the clusters.

\subsection{Semantic consistency}\label{sec:FRA-consistency}
In this section, we enforce the consistency of global embeddings and local facial embeddings. With the learned heatmaps $\mathbf{M}_i$, we generate the latent representations for the local facial regions through weighted average pooling:
\begin{align}\label{equ:FRA-pool}
	\mathbf{h}^{m}_i & = \mathbf{M}^{(m)}_i \otimes \mathbf{F}_i \nonumber \\
	& = \frac{1}{\sum_{u,v}{\mathbf{M}_i[m,u,v]}}\sum_{u,v}{\mathbf{M}_i[m,u,v]\mathbf{F}_i[*,u,v]},
\end{align}
where $\otimes$ denotes channel-wise weighted average pooling, $\mathbf{M}^{(m)}_i$ is the $m$-th channel (heatmap) of $\mathbf{M}_i$ and $\mathbf{h}^{m}_i \in \mathbb{R}^C$ is the corresponding latent representation produced with $\mathbf{M}^{(m)}_i$. The facial embeddings $\{\mathbf{z}^m_1: \mathbf{z}^m_1 \in \mathbb{R}^D\}^N_{m=1}$ and $\{\mathbf{z}^m_2: \mathbf{z}^m_2 \in \mathbb{R}^D\}^N_{m=1}$ are obtained accordingly with the local projector $H^l_\theta$ and $H^l_\xi$:
\begin{equation}\label{eq:FRA-embd}
	\begin{array}{l}
		\mathbf{z}^{m}_1 = H^l_\theta(\mathbf{h}^{m}_1), \\ 
		\mathbf{z}^{m}_2 = H^l_\xi(\mathbf{h}^{m}_2).
	\end{array}
\end{equation}
We then match the global embeddings and local facial embeddings across views using the negative cosine similarity in BYOL~\cite{grill2020bootstrap}:
\begin{align}\label{eq:FRA-consist-asy}
	\mathcal{L}_\text{sim}(\mathbf{z}_1, \mathbf{z}_2) = -(& \lambda_\text{c} \times f_\text{s}(G^g_\theta(\mathbf{z}_1), \mathbf{z}_2) + \nonumber \\
	+ & (1-\lambda_\text{c}) \times \frac{1}{N}\sum_{m=1}^{N}f_\text{s}(G^l_\theta(\mathbf{z}^m_1), \mathbf{z}^m_2)),
\end{align}
where $f_\text{s}(\mathbf{u}, \mathbf{v})=\frac{\mathbf{u}^\top\mathbf{v}}{{\lVert\mathbf{u}\rVert}_2{\lVert\mathbf{v}\rVert}_2}$ denotes the cosine similarity between the vectors $\mathbf{u}$ and $\mathbf{v}$, $\lambda_\text{c}$ is the loss weight, $G^g_\theta$ and $G^l_\theta$ are the predictors on top of the projectors $H^g_\theta$ and $H^l_\theta$, respectively. Following BYOL~\cite{grill2020bootstrap}, we symmetrize the loss $\mathcal{L}_\text{sim}(\mathbf{z}_1, \mathbf{z}_2)$ defined in~\cref{eq:FRA-consist-asy} by passing $\mathbf{x}_1$ through the momentum network $\xi$ and $\mathbf{x}_2$ through the online network $\theta$ to compute $\mathcal{L}_\text{sim}(\mathbf{z}_2, \mathbf{z}_1)$. The semantic consistency objective can be expressed as follows:
\begin{equation}\label{eq:FRA-consist} 
	\mathcal{L}_\text{c} = \mathcal{L}_\text{sim}(\mathbf{z}_1, \mathbf{z}_2) + \mathcal{L}_\text{sim}(\mathbf{z}_2, \mathbf{z}_1).
\end{equation}

\subsection{Overall objective}
We jointly optimize the semantic relation objective~\cref{eq:FRA-relation} and the semantic consistency objective~\cref{eq:FRA-consist}, leading to the following overall objective:
\begin{equation}\label{eq:scd-overall}
	\mathcal{L} = \mathcal{L}_\text{c} + \lambda_\text{r}\mathcal{L}_\text{r},
\end{equation}
where $\lambda_\text{r}$ is the loss weight for balancing $\mathcal{L}_\text{c}$ and $\mathcal{L}_\text{r}$.

\setlength{\tabcolsep}{4.5pt}
\begin{table*}[htb]
	\caption{\textbf{Comparisons with weakly-supervised pre-trained vision transformer on several downstream facial analysis tasks}, including facial expression recognition (AffectNet), facial attribute recognition (CelebA) and face alignment (300W).}
	\centering
	\label{tab:FRA-vt}
	\begin{tabular}{l c c c c c c c c}
		\toprule
		\multirow{2}{*}{Method} & \multirow{2}{*}{Arch.} & \multirow{2}{*}{Params.} & \multicolumn{3}{c}{Pre-training settings} & \multicolumn{3}{c}{Downstream performances} \\
		\cmidrule(lr){4-6} \cmidrule(lr){7-9}
		& & & Dataset & Scale & Supervision & \thead{AffectNet\\ Acc. $\uparrow$} & \thead{CelebA\\ Acc. $\uparrow$} & \thead{300W \\NME $\downarrow$} \\
		\midrule
		FaRL~\cite{zheng2022general} & ViT-B/16~\cite{dosovitskiy2021an} & 86M & LAION-FACE~\cite{zheng2022general} & 20M & face image + text & 64.85 & 91.88 & 3.08 \\
		\rowcolor{WhiteSmoke!70!Lavender} FRA & R50~\cite{he2016deep} & 24M & VGGFace2~\cite{cao2018vggface2} & 3.3M & face image & \textbf{66.16} & \textbf{92.02} & \textbf{2.91} \\
		\bottomrule
	\end{tabular}
\end{table*}
\setlength{\tabcolsep}{6pt}

\setlength{\tabcolsep}{4pt}
\begin{table}[htb]
	\caption{\textbf{Comparisons on facial expression recognition}. We report the Top-1 accuracy on test set. \textbf{Text} denotes text supervision. $^\dag$: our reproduction using the official codes.}
	\centering
	\label{tab:FRA-FER}
	\begin{tabular}{l c c c c}
		\toprule
		Method & Text & FERPlus & RAF-DB & AffectNet \\
		\midrule
		\multicolumn{5}{l}{\textbf{Supervised}} \\
		KTN~\cite{li2021adaptively} & \ding{53} & 90.49 & 88.07 & 63.97 \\
		RUL~\cite{zhang2021relative} & \ding{53} & 88.75 & 88.98 & 61.43 \\
		EAC~\cite{zhang2022learn} & \ding{53} & 90.05 & 90.35 & 65.32 \\
		\midrule
		\multicolumn{5}{l}{\textbf{Weakly-Supervised}} \\
		FaRL~\cite{zheng2022general}$^\dag$ & \checkmark & 88.62 & 88.31 & 64.85 \\
		CLEF~\cite{zhang2023weakly} & \checkmark & 89.74 & 90.09 & 65.66 \\
		\midrule
		\multicolumn{5}{l}{\textbf{Self-supervised}} \\
		MCF~\cite{wang2023toward}$^\dag$ & \ding{53} & 88.17 & 86.86 & 60.98 \\
		Bulat \etal~\cite{caron2020unsupervised,bulat2022pre} & \ding{53} & - & - & 60.20 \\
		BYOL~\cite{grill2020bootstrap} & \ding{53} & 89.25 & 89.53 & 65.65 \\
		LEWEL~\cite{huang2022learning} & \ding{53} & 85.61 & 81.85 & 61.20 \\
		PCL~\cite{liu2023pose} & \ding{53} & 85.87 & 85.92 & 60.77 \\
		\rowcolor{WhiteSmoke!70!Lavender} FRA (LP) & \ding{53} & 78.13 & 73.89 & 57.38 \\
		\rowcolor{WhiteSmoke!70!Lavender} FRA (FT) & \ding{53} & 89.78 & 89.95 & \textbf{66.16} \\
		\rowcolor{WhiteSmoke!70!Lavender} FRA (EAC) & \ding{53} & \textbf{90.62} & \textbf{90.76} & 65.85 \\
		\bottomrule
	\end{tabular}
\end{table}
\setlength{\tabcolsep}{6pt}

\begin{table}[htb]
	\caption{\textbf{Comparisons on CelebA~\cite{liu2015deep} facial attribute recognition}. We report the averaged accuracy over all attributes. $^\dag$: our reproduction using the official codes. $\ast$: results cited from~\protect\cite{zheng2022general}.}
	\centering
	\label{tab:FRA-attribute}
	\begin{tabular}{l c}
		\toprule
		Method & Acc. ($\%$) \\
		\midrule
		\multicolumn{2}{l}{\textbf{Supervised}} \\
		DMM~\cite{mao2020deep} & 91.70 \\
		SlimCNN~\cite{sharma2020slim} & 91.24 \\
		AFFAIR~\cite{li2018landmark} & 91.45 \\
		\midrule
		\multicolumn{2}{l}{\textbf{Self-supervised}} \\
		SSPL~\cite{shu2021learning} & 91.77 \\
		Bulat \etal~\cite{caron2020unsupervised,bulat2022pre}$\ast$ & 89.65 \\
		SimCLR~\cite{pmlr-v119-chen20j}$\ast$ & 91.08 \\
		BYOL~\cite{grill2020bootstrap} & 91.56 \\
		LEWEL~\cite{huang2022learning} & 90.69 \\
		PCL~\cite{liu2023pose} & 91.48 \\
		MCF~\cite{wang2023toward}$^\dag$ & 91.33 \\
		\rowcolor{WhiteSmoke!70!Lavender} FRA (LP) & 90.86 \\
		\rowcolor{WhiteSmoke!70!Lavender} FRA (FT) & \textbf{92.02} \\
		\bottomrule
	\end{tabular}
\end{table}

\setlength{\tabcolsep}{5pt}
\begin{table*}[htb]
	\caption{\textbf{Comparisons on face alignment}. $^\dag$: our reproduction using the official codes.}
	\centering
	\label{tab:FRA-landmark}
	\begin{tabular}{l c c c c c c c c}
		\toprule
		\multirow{2}{*}{Method} & Venue & \multirow{2}{*}{Arch.} & \multicolumn{3}{c}{WFLW} & \multicolumn{3}{c}{300W (NME $\downarrow$)} \\
		\cmidrule(lr){4-6} \cmidrule(lr){7-9}
		& & & NME $\downarrow$ & $\text{FR}_{10\%}$ $\downarrow$ & $\text{AUC}_{10\%}$ $\uparrow$ & Full & Comm. & Chal. \\
		\midrule
		\multicolumn{9}{l}{\textbf{Supervised}} \\
		SLPT~\cite{xia2022sparse} & [CVPR’22] & ResNet~\cite{he2016deep} & 4.20 & 3.04 & 0.588 & 3.20 & 2.78 & 4.93 \\
		DTLD~\cite{li2022towards} & [CVPR’22] & ResNet~\cite{he2016deep} & 4.08 & 2.76 & - & 2.96 & 2.59 & 4.50 \\
		RePFormer~\cite{li2022repformer} & [IJCAI’22] & ResNet~\cite{he2016deep} & 4.11 & - & - & 3.01 & - & - \\
		ADNet~\cite{huang2021adnet} & [ICCV’21] & Hourglass~\cite{yang2017stacked} & 4.14 & 2.72 & 0.602 & 2.93 & 2.53 & 4.58 \\
		STAR~\cite{zhou2023star} & [CVPR’23] & Hourglass~\cite{yang2017stacked} & 4.02 & 2.32 & 0.605 & 2.87 & 2.52 & 4.32 \\
		\midrule
		\multicolumn{9}{l}{\textbf{Self-supervised}} \\
		MCF~\cite{wang2023toward} (concurrent work) & [ACM MM’23] & ViT~\cite{dosovitskiy2021an} & \textbf{3.96} & \textbf{1.40} & \textbf{0.609} & 2.98 & \textbf{2.60} & 4.51 \\
		Bulat \etal~\cite{caron2020unsupervised,bulat2022pre} & [ECCV’22] & ResNet~\cite{he2016deep} & 4.57 & - & - & 3.20 & - & - \\
		BYOL~\cite{grill2020bootstrap} & [NeurIPS’20] & ResNet~\cite{he2016deep} & 4.29 & 2.96 & 0.579 & 3.03 & 2.66 & 4.55 \\
		LEWEL~\cite{huang2022learning} & [CVPR’22] & ResNet~\cite{he2016deep} & 4.52 & 4.50 & 0.563 & 3.09 & 2.70 & 4.71 \\
		PCL~\cite{liu2023pose}$^\dag$ & [CVPR’23] & ResNet~\cite{he2016deep} & 4.84 & 6.18 & 0.535 & 3.35 & 2.77 & 5.12 \\
		\rowcolor{WhiteSmoke!70!Lavender} FRA & Ours & ResNet~\cite{he2016deep} & 4.11 & 2.53 & 0.591 & \textbf{2.91} & \textbf{2.60} & \textbf{4.46} \\
		\bottomrule
	\end{tabular}
\end{table*}
\setlength{\tabcolsep}{6pt}

\section{Experiments}

\subsection{Experimental setups}
\subsubsection{Implementation details}
We use the same augmentation strategy as in~\cite{grill2020bootstrap,huang2022learning}. The number of heatmaps $N$ is set to 8 empirically. The loss weight $\lambda_\text{c}$ and $\lambda_\text{r}$ are set to 0.5/0.1, respectively. For fair comparisons, the other hyper-parameters are kept the same as BYOL~\cite{grill2020bootstrap} in all experiments. The architecture and pre-training details are provided in the supplementary material.

\subsubsection{Baselines}
Our baselines are self-supervised pre-training approaches for visual images (\eg, BYOL~\cite{grill2020bootstrap} and LEWEL~\cite{huang2022learning}), and pre-training approaches for facial images (\eg, Bulat \etal~\cite{bulat2022pre} and PCL~\cite{liu2023pose}). Note that SwAV~\cite{caron2020unsupervised} is equivalent to Bulat \etal~\cite{bulat2022pre}. As we adopt BYOL~\cite{grill2020bootstrap} as the pre-training backbone, we compare our FRA with BYOL~\cite{grill2020bootstrap} in all experiments. We also compare our FRA with another pre-training method LEWEL~\cite{huang2022learning}, which learns consistent local representations for visual images. Moreover, we perform comparisons with SOTA methods in the corresponding downstream tasks.

\subsection{Evaluation protocols}\label{sec:FRA-evaluation}
Following the common practice in previous works~\cite{zheng2022general,liu2023pose}, we evaluate the transfer performance of the self-supervised pre-trained facial representations on several popular downstream facial analysis tasks: facial expression recognition (FER)~\cite{barsoum2016training,li2017reliable}, facial attribute recognition (FAR)~\cite{liu2015deep} and face alignment (FA)~\cite{wu2018look,sagonas2016300,sagonas2013300,sagonas2013semi}. Specifically, we use the pre-trained weights to initialize the backbone of downstream tasks and then learn the backbone and task-specific head (attached to the backbone) jointly. Following~\cite{zheng2022general}, we report the performance with linear probe (denoted by ``\textbf{LP}'') and fine-tuning (denoted by ``\textbf{FT}''). The details of the downstream tasks are described as follows:

\textbf{Facial expression recognition} is a multi-class classification task where the goal is to categorize the emotional expressions (\eg, anger, fear and surprise) for a given face image. Three widely-used datasets are adopted: FERPlus~\cite{barsoum2016training}, RAF-DB~\cite{li2017reliable} and AffectNet~\cite{mollahosseini2017affectnet}. For RAF-DB, we use the basic emotion subset following~\cite{li2021adaptively,zhang2022learn,liu2023pose}. For AffectNet, we report the results with 7 emotion classes (\ie, neutral, happy, sad, surprise, fear, anger, disgust) following~\cite{li2021adaptively,zhang2022learn}.

\textbf{Facial attribute recognition} is a multi-label classification task to predict various attributes (\eg, gender, age and race) of a given face image. We adopt the popular benchmark CelebA~\cite{liu2015deep}, which consists of more than 200K face images with 40 facial attributes per image. Following~\cite{zheng2022general}, we report the averaged accuracy over all attributes.

\textbf{Face alignment} is a regression task to predict 2D face landmark coordinates on a face image. We use two popular benchmarks: WFLW~\cite{wu2018look} and 300W~\cite{sagonas2016300,sagonas2013300,sagonas2013semi}. Following the common practice~\cite{cheng2021equivariant,huang2021adnet,zhou2023star}, we report normalized mean error (NME), failure rate (FR) and AUC. For 300W, we report the results on full test set, common (554 images) and challenge (135 images) splits of the test set following~\cite{huang2021adnet,zhou2023star}.

\subsection{Comparisons with weakly-supervised pre-training}
In~\cref{tab:FRA-vt}, we compare our FRA with SOTA pre-trained Transformer FaRL~\cite{zheng2022general}, which is a weakly-supervised model pre-trained on 20M visual-linguistic data (face image and text) with image-text contrastive learning and mask image modeling. We fine-tune both the pre-trained feature backbone and the task-specific head on the corresponding downstream facial analysis task. Our self-supervised FRA with 24M parameter ResNet-50 achieves superior performance compared with weakly-supervised FaRL~\cite{zheng2022general} with 86M parameter ViT-B/16 and text supervision on all tasks.

\subsection{Transfer learning}
In this section, we compare our FRA with self-supervised pre-training approaches and SOTA methods in several downstream tasks. Please refer to the supplementary material for setup details.

\subsubsection{Facial expression recognition}
The results on facial expression recognition are reported in~\cref{tab:FRA-FER}. We observe: \textbf{(1)} With the setting of fine-tuning (FT), our FRA outperforms previous self-supervised pre-training approaches for visual images (\eg, BYOL and LEWEL) and pre-training approaches tailored for facial images (\eg, PCL, MCF). In particular, our FRA using a 24M parameter ResNet-50 surpasses the concurrent work MCF~\cite{wang2023toward} with 86M parameter ViT-B/16~\cite{dosovitskiy2021an}. \textbf{(2)} By simply learning a linear classifier on top of the encoder backbone, our FRA outperforms SOTA facial expression recognition methods with sophisticated designs (\eg, EAC~\cite{zhang2022learn}) on AffectNet~\cite{mollahosseini2017affectnet}, the largest facial expression recognition dataset. \textbf{(3)} More importantly, by using our pre-trained model to initialize the backbone of SOTA facial expression recognition method EAC~\cite{zhang2022learn}, our variant ``FRA (EAC)'' consistently improves EAC~\cite{zhang2022learn} on all datasets, which suggests ``FRA (EAC)'' outperforms SOTA FER methods and demonstrates the superiority of the proposed self-supervised pre-training.

\subsubsection{Facial attribute recognition}
As shown in~\cref{tab:FRA-attribute}, our FRA outperforms both self-supervised pre-training approaches for visual images and pre-training approaches tailored for facial images. The results on facial expression recognition and facial attribute recognition show that our FRA learns better facial representations for facial classification task.

\subsubsection{Face alignment}
As shown in~\cref{tab:FRA-landmark}, despite SOTA face alignment methods (\eg, ADNet~\cite{huang2021adnet} and STAR~\cite{zhou2023star}) commonly rely on hourglass network~\cite{yang2017stacked} for feature extraction, which is tailored for regression tasks like landmark detection, our method based on ResNet backbone achieves comparable performance with these SOTA methods (\eg, 2.91 vs. 2.87 on 300W). The results on classification (\eg, facial expression recognition) and regression tasks (\eg, face alignment) show that \textbf{our FRA achieves SOTA results using vanilla ResNet~\cite{he2016deep} as the unified backbone for various facial analysis tasks}.

\begin{table}[htb]
	\caption{\textbf{Effect of different modules}. \textbf{GC} denotes the global consistency for aligning images, \textbf{LC} denotes the local consistency for aligning facial regions with heatmaps and \textbf{SR} represents semantic relation for aligning pixels and heatmaps.}
	\centering
	\label{tab:FRA-modules}
	\begin{tabular}{c c c c c c}
		\toprule
		GC & LC & SR & RAF-DB $\uparrow$ & CelebA $\uparrow$ & 300W $\downarrow$ \\
		\midrule
		\checkmark & - & - & 87.82 & 90.63 & 3.56 \\
		- & - & \checkmark & 85.34 & 90.66 & 3.25 \\
		- & \checkmark & - & 87.46 & 90.78 & 3.19 \\
		\checkmark & \checkmark & - & 88.05 & 90.89 & 3.26 \\
		\checkmark & \checkmark & \checkmark & \textbf{88.72} & \textbf{91.18} & \textbf{3.14} \\
		\bottomrule
	\end{tabular}
\end{table}

\begin{table}[htb]
	\caption{Effect of the number of heatmaps $N$.}
	\centering
	\label{tab:FRA-heatmap}
	\begin{tabular}{l c c c}
		\toprule
		$N$ & 8 & 32 & 64 \\
		\midrule
		RAF-DB & \textbf{88.72} & 88.36 & 88.18 \\
		CelebA & \textbf{91.18} & 91.01 & 90.98 \\
		\bottomrule
	\end{tabular}
\end{table}

\setlength{\tabcolsep}{5pt}
\begin{table}[htb]
	\caption{\textbf{Effect of the loss weights}. Please refer to~\cref{tab:FRA-modules} for \textbf{GC}, \textbf{LC} and \textbf{SR}.}
	\centering
	\label{tab:FRA-weight}
	\begin{tabular}{l c c c c}
		\toprule
		Settings & $\lambda_\text{c}$ & $\lambda_\text{r}$ & RAF-DB & CelebA \\
		\midrule
		GC & 1.0 & 0 & 87.82 & 90.63 \\
		GC + LC & 0.5 & 0 & 88.05 & 90.89 \\
		GC + LC + SR (FRA) & 0.5 & 0.1 & \textbf{88.72} & \textbf{91.18} \\
		GC + LC + SR (FRA) & 0.5 & 0.5 & 88.45 & 91.04 \\
		GC + LC + SR (FRA) & 0.5 & 1.0 & 88.08 & 90.46 \\
		\bottomrule
	\end{tabular}
\end{table}
\setlength{\tabcolsep}{6pt}

\subsection{Ablation studies}
We pre-train the model on VGGFace2~\cite{cao2018vggface2} and then evaluate it on facial expression recognition (RAF-DB) and facial attribute recognition (CelebA), as described in~\cref{sec:FRA-evaluation}.

\subsubsection{Effect of different modules}
In~\cref{tab:FRA-modules}, we investigate the contributions of the proposed semantic consistency loss (i.e, global consistency of whole face and local consistency of facial regions) and semantic relation loss to our approach. Note that the global consistency (first row) is BYOL~\cite{grill2020bootstrap}. We have the following observations: \textbf{(1)} The variant using all losses achieves the best results. \textbf{(2)} GC is essential to avoid degeneration on classification task. \textbf{(3)} LC or SR alone benefits regression task (landmark). Altogether, LC and SR improve BYOL~\cite{grill2020bootstrap} (GC) on both classification and regression by capturing spatial/local information, which validates our facial region awareness.

\subsubsection{Effect of the number of heatmaps}
In~\cref{tab:FRA-heatmap}, we study the effect of the number of heatmaps. We observe that the best setting is $N=8$, which is close to the facial landmarks number 5. This suggests that given enough face images for training, a suitable $N$ can encourage the model to learn face-specific patterns, which helps the transfer learning performance on various facial analysis tasks. Further increasing the number of heatmaps might force the model to look into fine-grained patterns that may not be suitable for facial tasks.

\subsubsection{Effect of loss weights}
In~\cref{tab:FRA-weight}, we ablate the weights for the semantic consistency loss and semantic relation loss. We find that setting $\lambda_\text{c}=0.5$ and $\lambda_\text{r}=0.1$ works best. When $\lambda_\text{c}=1.0$ and $\lambda_\text{r}=0$, only the consistency of global representations is applied, and the model performs relatively worse, which suggests the importance of the consistency of local representations and the semantic relation loss. By using the semantic relation objective, the performance is significantly improved. However, when $\lambda_\text{r}$ is too high, the performance degrades as the pixel-level consistency between the online and momentum network might affect the capture of image/object-level information.

\subsubsection{Effect of Transformer decoder layers}
\begin{table}[htb]
	\caption{\textbf{Effect of Transformer decoder layers}. 0 decoder layer represents BYOL~\cite{grill2020bootstrap} where only the consistency of global representation is enforced.}
	\centering
	\label{tab:FRA-dec}
	\begin{tabular}{l c c c c}
		\toprule
		\# decoder layer & 0 & 1 & 2 & 3 \\
		\midrule
		RAF-DB & 87.82 & 88.72 & 89.01 & 89.06 \\
		CelebA & 90.63 & 91.18 & 91.30 & 91.35 \\
		\bottomrule
	\end{tabular}
\end{table}

In~\cref{tab:FRA-dec}, we study the effect of the number of decoder layers used for heatmap prediction. We observe that a single decoder layer is able to produce decent results, showing that a 1-layer decoder is large enough to capture the facial region (landmarks) relations in face images. The performance gain diminishes as the decoder depth increases. By default, we only use 1 decoder layer for fast training.

\section{Conclusion}
In this work, we propose a novel self-supervised facial representation learning framework to learn consistent global and local facial representations, \textbf{F}acial \textbf{R}egion \textbf{A}wareness (FRA). We learn a set of heatmaps indicating facial regions from learnable positional embeddings, which leverages the attention mechanism to look up facial image globally for facial regions. We show that our FRA outperforms previous pre-trained models on several facial classification and regression tasks. More importantly, using ResNet as the unified backbone, our FRA achieves comparable or even better performance compared with SOTA methods in facial analysis tasks.

\section*{Acknowledgement}
This work was supported by the EU H2020 AI4Media No.951911 project. We thank Zengqun Zhao for his helpful comments on facial expression recognition.

{
    \small
    \bibliographystyle{ieeenat_fullname}
    \bibliography{references}
}


\end{document}